\documentclass{article} 
\usepackage{iclr2015_conference,times}
\usepackage{hyperref}
\usepackage{url}

\usepackage{amsfonts}
\usepackage{amssymb}
\usepackage{amsmath}
\usepackage{geometry}

\usepackage{graphicx}
\usepackage{float}
\usepackage{booktabs}
\usepackage[table,xcdraw,usenames,dvipsnames]{xcolor}
\usepackage{color}
\usepackage{dcolumn}
\newcolumntype{d}[1]{D{.}{.}{#1}}
\usepackage{array}
\newcommand{\head}[2]{\multicolumn{1}{>{\centering\arraybackslash}p{#1}|}{{#2}}}
\usepackage{epstopdf}
\usepackage{caption}
\usepackage{subcaption}

\usepackage{multirow}

\definecolor{darkgreen}{rgb}{0.0, 0.4, 0.0}
\definecolor{brown(web)}{rgb}{0.65, 0.16, 0.16}
\title{Sequence to Sequence Learning for Optical Character Recognition}

\author{Devendra Kumar Sahu \&  Mohak Sukhwani\\
International Institute of Information Technology\\
Hyderabad, Telangana 500032, India \\
\texttt{dksahuji@gmail.com} \\ \texttt{mohak.sukhwani@research.iiit.ac.in}\\
}

\begin{document}

\maketitle
\begin{abstract}
We propose an end-to-end recurrent encoder-decoder based sequence learning approach for printed text Optical Character Recognition (\textsc{ocr}). In contrast to present day existing state-of-art \textsc{ocr} solution [\cite{Graves:2006:CTC:1143844.1143891}] which uses \textsc{ctc} output layer, our approach makes minimalistic assumptions on the structure and length of the sequence. We use a two step encoder-decoder approach -- (a) A recurrent encoder reads a variable length printed text word image and encodes it to a fixed dimensional embedding. (b) This fixed dimensional embedding is subsequently comprehended by decoder structure which converts it into a variable length text output. 
Our architecture gives competitive performance relative to Connectionist Temporal Classification (\textsc{ctc}) [\cite{Graves:2006:CTC:1143844.1143891}] output layer while being executed in more natural settings. The learnt deep word image embedding from encoder can be used for printed text based retrieval systems. The expressive \textit{fixed} dimensional embedding for any variable length input expedites the task of retrieval and makes it more \textit{efficient} which is not possible with other recurrent neural network architectures. We empirically investigate the expressiveness and the learnability of long short term memory (\textsc{lstm}s) in the sequence to sequence learning regime by training our network for prediction tasks in segmentation free printed text {\textsc{ocr}}s. The utility of the proposed architecture for printed text is demonstrated by quantitative and qualitative evaluation of two tasks -- word prediction and retrieval. 
\end{abstract}
\section{Introduction}
Deep Neural Nets (\textsc{dnn}s) have become present day de-facto standard for any modern 
machine learning task. The flexibility and power of such structures have made them 
outperform other methods in solving some really complex problems of speech [\cite{deepSpeechReviewSPM2012}] and 
object [\cite{Krizhevsky_imagenetclassification}] recognition. We exploit the power of such structures in an \textsc{ocr} 
based application for word prediction and retrieval with a single model. Optical character recognition (\textsc{ocr}) is the task of converting images of typed, handwritten or printed text into machine-encoded text. It is a  method of 
digitizing printed texts so that it can be electronically edited, searched, stored 
more compactly, displayed on-line and used in machine processes such as machine 
translation, text-to-speech and text mining. 

\begin{figure}
\begin{subfigure}{.5\textwidth}
  \centering
  \includegraphics[height=5cm]{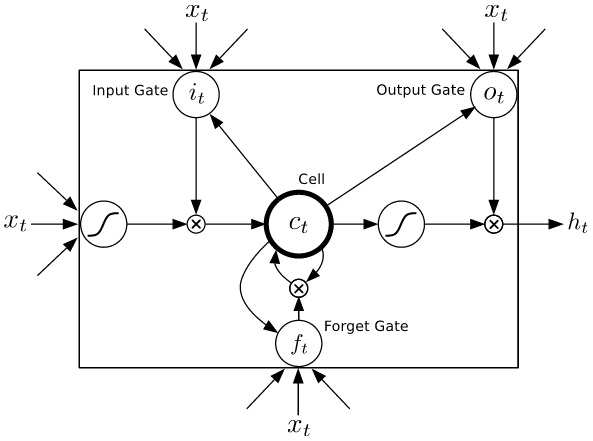}
 \caption{LSTM block}
\label{fig:LSTM}
\end{subfigure}%
\begin{subfigure}{.5\textwidth}
  \centering
  \includegraphics[height=5cm]{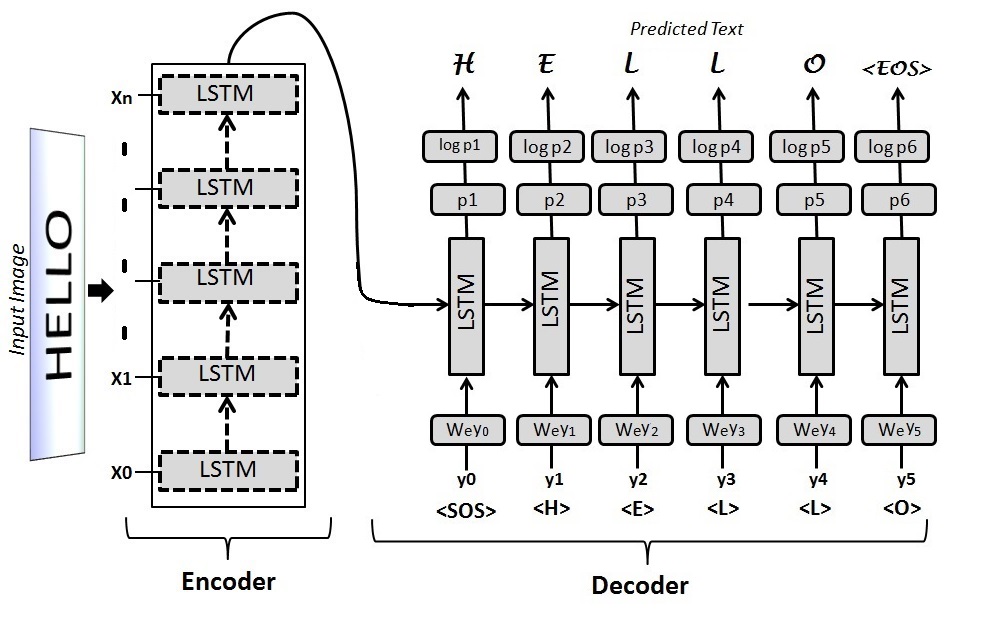}
\caption{Encoder Decoder Framerwork for OCR}
\label{fig:lstmED}
\end{subfigure}
\caption{Figure \ref{fig:LSTM} is the \textsc{lstm} block comprised of input, output and forget gates. Figure \ref{fig:lstmED} showcases the proposed recurrent encoder decoder framework for Optical Character Recognition. The Encoder \textsc{lstm} section reads the input image and converts it to a fixed-dimensional vector representation. Decoder \textsc{lstm} in turn generates the text output corresponding to fixed-dimensional vector representation.}
\end{figure}

From character recognition to word prediction, \textsc{ocr}s in recent years have 
gained much awaited traction in mainstream applications. With its usage spanning 
across handwriting recognition, print text identification, language 
identification etc.~\textsc{ocr}s have humongous untapped potential. In our 
present work we show an end-to-end, deep neural net, based architecture 
for word prediction and retrieval. We conceptualize the problem as that of a 
sequence to sequence learning and use \textsc{rnn} based architecture to 
first encode input to a fixed dimension feature and later decode it to variable 
length output. Recurrent Neural Networks (\textsc{rnn}) architecture has an innate 
ability to learn data with sequential or temporal structure. This makes them suitable 
for our application. Encoder \textsc{lstm} network reads the input sequence one step at a time and 
converts it to an expressive fixed-dimensional vector representation. 
Decoder \textsc{lstm} network in turn converts this fixed-dimensional vector (Figure \ref{fig:lstmED}) to the text output.

Encoder-Decoder framework has been applied to many applications recently. 
[\cite{Sutskever_generatingtext}] used recurrent encoder-decoder for character-level language 
modelling task where they predict the next character given the past predictions. 
It has also been used for language translation [\cite{SutskeverVL14_Seq2Seq}] 
where a complete sentence is given as input in one language and the decoder 
predicts a complete sentence in another language. 
Vinyals et al. [\cite{VinyalsTBE14_captionGeneration}] presented a model based on a deep recurrent architecture 
to generate natural sentences 
describing an image. They used a convolutional neural network as encoder and 
a recurrent decoder to describe images in natural language. 
Zaremba et al. [\cite{ZarembaS14_Learn2Execute}] used sequence to sequence learning for evaluating short computer programs, a domain that have been seen as too complex in past. 
Vinyals et al. [\cite{VinyalsL15_ConversationalNN}] proposed neural conversational 
networks based of sequence to sequence learning framework which converses by predicting the next sentence given the previous sentence(s) in a conversation. In the same spirit as \cite{VinyalsTBE14_captionGeneration,SutskeverVL14_Seq2Seq}, we formulate the \textsc{ocr} problem as a sequence to sequence mapping problem to convert an input (text) image to its corresponding text.

In this paper, we investigate the expressiveness and learnability of \textsc{lstm}s in sequence to 
sequence learning regime for printed text \textsc{ocr}. We demonstrate that sequence to sequence
learning is suitable for word prediction task in a segmentation free setting. We even show the 
expressiveness of the learnt deep word image embeddings (from Encoder network of prediction) on image retrieval task. 
In (majority of) cases where standard \textsc{lstm} models do not convert a variable length input to a fixed dimensional output, we are required to use Dynamic Time Warping (\textsc{dtw}) for retrieval which tends to be computationally expensive and slow. Converting variable length samples to fixed dimensional representation gives us access to fast and efficient methods for retrieval in fixed dimensional regime –- approximate nearest neighbour.
 
\section{Sequence learning}
\label{sec:Model}

A recurrent neural network (\textsc{rnn}) is a neural network with cyclic connections between its units. These cycles create a concept of `internal memory' in network and thus differentiate \textsc{rnn}s from other feed forward networks. The internal memory of \textsc{rnn} can be used to process arbitrary sequences of inputs -- given a variable length input sequence $X$ we can generate corresponding variable length output sequence $Y$. This is done by sequentially reading each time-step $x_t$ of input sequence $X$ and updating its internal hidden representations $h_t$. More sophisticated recurrent activation functions like \textsc{lstm} [\cite{Hochreiter:1997:LSM:1246443.1246450}] and \textsc{gru} [\cite{GRU_ChoMGBSB14, GRU_ChungGCB14}] have become more common in recent days. They perform better when compared to other vanilla \textsc{rnn} implementations.

Long Short-Term Memory [\cite{Hochreiter:1997:LSM:1246443.1246450}] is a \textsc{rnn} 
architecture that elegantly addresses the vanishing gradients problem using `memory units'. 
These linear units have a pair of auxiliary `gating units' that control the flow of information 
to and from the unit. Equations \ref{eq:LSTM_it}-\ref{eq:LSTM_ht} describe \textsc{lstm} blocks.
\begin{align} 
\label{eq:LSTM_it}
 i_t &= \sigma(W_{xi}x_t + W_{hi}h_{t-1} + w_{ci}\odot c_{t-1} + b_i)\\
 \label{eq:LSTM_ft}
 f_t &= \sigma(W_{xf}x_t + W_{hf}h_{t-1} + w_{cf}\odot c_{t-1} + b_f) \\
 \label{eq:LSTM_ct}
 c_t &= f_{t}c_{t-1} + i_{t}tanh(W_{xc}x_t + W_{hc}h_{t-1} + b_c)\\
 \label{eq:LSTM_ot}
 o_t &= \sigma(W_{xo}x_t + W_{ho}h_{t-1} + w_{co}\odot c_{t-1} + b_o)\\
 \label{eq:LSTM_ht}
 h_t &= o_t tanh(c_t) 
\end{align}
Here, $W_{xi}, W_{hi}, w_{ci}, b_i$ are input gate parameters. $ W_{xo}, W_{ho}, w_{co}, b_o$ are output gate parameters. $ W_{xf}, W_{hf}, w_{cf}, b_f$ are forget gate parameters. $ W_{xc}, W_{hc}, b_c$ are parameters associated with input which directly modify the memory cells. The symbol $\odot$ denotes element-wise multiplication. The gating units are implemented by multiplication, so it is natural to restrict their domain to $[0, 1]^N$, which corresponds to the sigmoid non-linearity. The other units do not have this restriction, so the tanh non-linearity is more appropriate. We use collection of such 
units (Figure ~\ref{fig:LSTM}) to describe an encoder-decoder framework for the \textsc{ocr} task. We formulate the task of \textsc{ocr} prediction as a mapping problem between structured input (image) and structured output (text).

Let, $\{I_i, Y_i \}_{i=1}^{N}$ define our dataset with $I_i$ being image and $Y_i$ be the corresponding text label. Image $I_i$ lies in $\mathbb{R}^{H \times T_i}$, where $H$ is word image height (common for all images) and $T_i$ is the width of $i^{th}$ word image. We represent both image and label as a sequence -- Image $I_i = \{x_1, x_2, ..., x_{T_i}\} $ is sequence of $T_i$ vectors lying in $\{0,1\}^{H}$ ($x_i$ is a $i_{th}$ pixel column) and $Y_i = \{y_1, y_2, ..., y_{M_i}\}$ is corresponding label which is a sequence of $M_i$ unicode characters. We learn a mapping from image to text in two steps -- (a) $f_{encoder}: I_i \rightarrow z_i$, maps an image $I_i$ to a latent fixed dimensional representation $z_i$ (b) $f_{decoder}: z_i \rightarrow Y_i$ maps it to the output text sequence $Y_i$. Unlike \textsc{ctc} layer based sequence prediction [\cite{Graves:2006:CTC:1143844.1143891}] we don't have any constraint on length of sequences $I_i$ and $Y_i$. Equations \ref{eq:fEncDec1}-\ref{eq:fEncDec2} formally describe the idea. The choice of $f_{encoder}$ and $f_{decoder}$ depends on type of input and output respectively. Both input and output correspond to a sequence in our case, hence we use recurrent encoder and recurrent decoder formulation.
\begin{align} 
\label{eq:fEncDec1}
 z_i &= f_{encoder} (I_i)\\
 \label{eq:fEncDec2}
 P(Y_i | I_i) &= f_{decoder} (z_i)
  \end{align} 
\subsection{Encoder: \textsc{lstm} based Word Image Reader}
\label{subsec:Encoder}

To describe the formulation we use vanilla \textsc{rnn}s with $L$ hidden layer and no output layer. The encoder reads $I_i = \{x_1, x_2, ..., x_{T_i}\}$ one step at a time from $x_1$ to $x_{T_i}$. Hidden state $h^n_{t}$ is updated using equations \ref{eq:vanillaRNN_Encoder1}-\ref{eq:vanillaRNN_Encoder2} using current input $x_t$ and previous hidden state $\{h^n_{t-1}\}_{n=1}^{L}$ where $L$ is the number of hidden layers in \textsc{rnn}. 
\begin{align} 
\label{eq:vanillaRNN_Encoder1}
h^1_t &= relu(W_{ih^1}x_t + W_{h^1h^1}h^1_{t-1} + b_h^1)\\
\label{eq:vanillaRNN_Encoder2}
h^n_t &= relu(W_{h^{n-1}h^n}h^{n-1}_{t} + W_{h^{n}h^n}h^{n}_{t-1} + b_h^n)
\end{align}
where $W_{ih^1}, W_{h^1h^1}, b_h^1, W_{ih^n},  W_{h^{n-1}h^n}, W_{h^{n}h^n},  b_h^n$ are parameters to be learned.

To obtain our fixed dimensional latent representation $z_i$ we use the final hidden states $\{h^n_{T_i}\}_{n=1}^L$. 
\begin{align} \label{eq:z_i}
 z_i &= \{h^n_{T_i}\}_{n=1}^L
\end{align}
It should be noted that we have used \textsc{lstm} networks instead of vanilla \textsc{rnn}s and no output layer for encoder is needed. The hidden states of last step $T_i$ are used as initial state of decoder network. 

\subsection{Decoder: \textsc{lstm} based Word Predictor}
\label{subsec:Decoder}
Similar to encoder, we describe the idea using vanilla \textsc{rnn}s with $L$ hidden layers and softmax output layer. The goal of word predictor is to estimate the conditional probability $p(Y_i | I_i)$ as shown in equation \ref{eq:conditional_Decoder1}-\ref{eq:conditional_Decoder2}, where $I_i$ is image input sequence and $Y_i$ is output sequence. 
\begin{align} 
\label{eq:conditional_Decoder1}
p(Y_i | I_i) &= p(Y_i = \{y_1, \hdots, y_{T'} \}| I_i = \{x_1, \hdots, x_{T_i}\}) \\ 
\label{eq:conditional_Decoder2}
&= \prod_{t=1}^{T'} p(y_t | I_i, y_1, \hdots, y_{t-1}) 
\end{align}
The updates for single step for \textsc{rnn} is described in equations \ref{eq:vanillaRNN_Decoder1}-\ref{eq:vanillaRNN_Decoder4}. The hidden state $h^n_{t}$ is updated using equations  \ref{eq:vanillaRNN_Decoder1} - \ref{eq:vanillaRNN_Decoder2} using current input $x_t$ and previous hidden state $\{h^n_{t-1}\}_{n=1}^{L}$, where $L$ is number of hidden layers in \textsc{rnn}. The hidden activations, $h^L_{t-1}$ are used to predict the output at step $t$ using equations \ref{eq:vanillaRNN_Decoder3}-\ref{eq:vanillaRNN_Decoder4}. $x_t$ is the embedding of the most probable state in previous step $t-1$ shown in equation \ref{eq:embedding_Decoder}. 
\begin{align}
\label{eq:vanillaRNN_Decoder1}
 h^1_t &= relu(W_{ih^1}x_t + W_{h^1h^1}h^1_{t-1} + b_h^1)\\
 \label{eq:vanillaRNN_Decoder2}
  h^n_t &= relu(W_{h^{n-1}h^n}h^{n-1}_{t} + W_{h^{n}h^n}h^{n}_{t-1} + b_h^n)\\
  \label{eq:vanillaRNN_Decoder3}
  o_t &= W_{h^{L}o}h^{L}_{t} + b_o\\
  \label{eq:vanillaRNN_Decoder4}
  p_t &= sofmax(o_t)
\end{align}
where $W_{ih^1}, W_{h^1h^1}, b_h^1, W_{ih^n},  W_{h^{n-1}h^n}, W_{h^{n}h^n},  b_h^n, W_{h^{L}o}, b_o$ are parameters to be learned.

Decoding begins at $t=0$ with $\langle SOS \rangle$ marker (Start Of Sequence). The state $t=-1$ is initialized with the final state $z_i$ of encoder as shown in equation \ref{eq:init_Decoder}. The $1^{st}$ character is predicted using the embedding for $\tilde{y}_{0} = \langle SOS \rangle$ as input $x_{0} = W_e \tilde{y}_{0}$ and output $p_1(y_{1} | I_i, y_0)$ where $W_e$ is the embedding matrix for characters. As shown in equation \ref{eq:embedding_Decoder}, every ${t_{th}}$ character is predicted using the embedding for $\tilde{y}_{t} = {\arg \max}~p_{t}(y_{t} | I_i, y_{< t})$ as input and output $p_{t}(y_{t} | I_i, y_{< t})$. This is iterated till $t=T'$ where $y_{T'} = \langle  EOS \rangle$ (End Of Sequence). $T'$ is not known in priori, $\langle  EOS \rangle$ marker instantiates the value of $T'$.
\begin{align} 
\label{eq:init_Decoder}
h_{-1} &= f_{encoder} (I) \\
\label{eq:embedding_Decoder}
x_t &= W_e \tilde{Y}_t \\
\label{eq:prob_Decoder}
p_{t+1} &= f_{decoder}(x_t)
\end{align}
$W_e$ in equation \ref{eq:embedding_Decoder} is the embedding matrix for characters.
It should be (again) noted that we use \textsc{lstm} networks (equations \ref{eq:LSTM_it} - \ref{eq:LSTM_ht}) instead of vanilla \textsc{rnn}s (equations \ref{eq:vanillaRNN_Decoder1}, \ref{eq:vanillaRNN_Decoder2}). 
\subsection{Training}
\label{subsec:Training}
The model described in section \ref{subsec:Encoder} and \ref{subsec:Decoder} is trained to predict characters of the input word (image) sequence. The input at time $t$ of decoder is an embedding of the output of time $t-1$. The loss $L$ for a sample $(I,Y)$ is described by equation \ref{eq:neglogLikelihood}. 
\begin{align} \label{eq:neglogLikelihood}
L(I,Y) &= - \log p(Y | I ; \theta)
\end{align}
Here, $\log p(Y | I ; \theta) = \sum_{t=0}^M \log p(y_t | I, y_0, \ldots, y_{t-1} ; \theta)$ is the $\log$ probability of correct symbol at each step. 
We search for the parameters $\theta^*$ which minimize the expected loss over true distribution $P(I,Y)$ given in equation \ref{eq:loss_optimization}. This distribution is unknown and can be approximated with empirical distribution $\tilde{P}(I,Y)$ given in equation \ref{eq:approx_loss_optimization}-\ref{eq:final_loss_optimization}. 
\begin{align} 
\label{eq:loss_optimization}
\theta^* &= \underset{\theta}{\arg\min} ~\mathbb{E}_{P(I,Y)} L(I,Y; \theta) \\
\label{eq:approx_loss_optimization}
&\approx \underset{\theta}{\arg\min} ~\mathbb{E}_{\tilde{P}(I,Y)} L(I,Y; \theta) \\
\label{eq:final_loss_optimization}
&= \underset{\theta}{\arg\min} \frac{1}{N} \sum_{i=1}^N L(I_i,Y_i; \theta) 
\end{align}

\begin{figure}
\begin{subfigure}{.48\textwidth}
  \centering
  \includegraphics[width=1\linewidth, trim=65mm 28mm 50mm 30mm, clip]{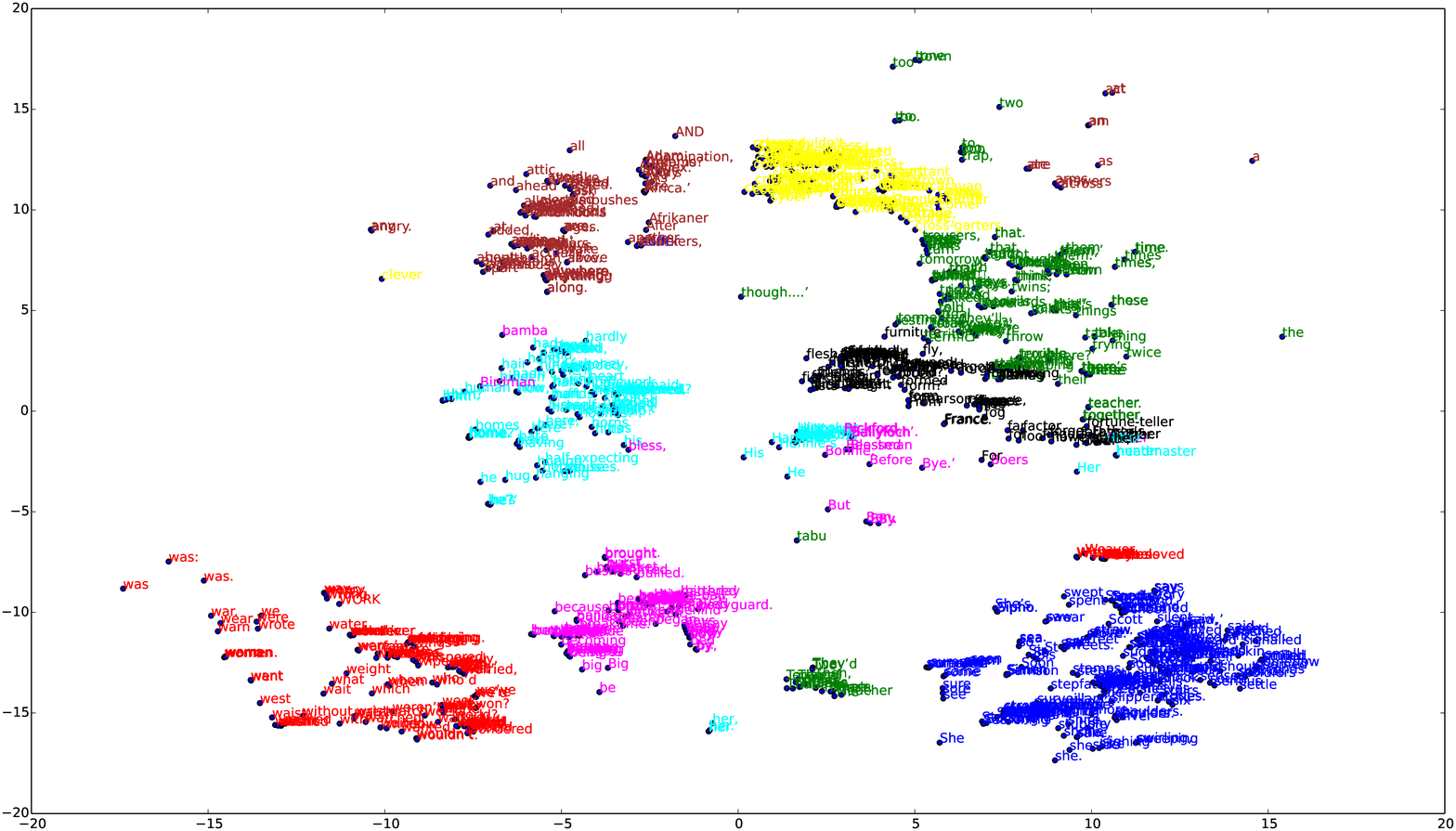}
 \caption{TSNE plots for feature representation of unique words starting with top eight most frequent alphabets({\color{blue}S}, {\color{darkgreen}T}, {\color{red}W}, {\color{cyan}H}, {\color{magenta}B}, {\color{yellow}C}, {F}, {\color{brown(web)}A} including lowercase. (Color based on first alphabet of words)}
\label{fig:WordVizLevel1}
\end{subfigure}%
\hfill
\begin{subfigure}{.48\textwidth}
  \centering
  \includegraphics[width=1\linewidth, trim=65mm 28mm 50mm 30mm, clip]{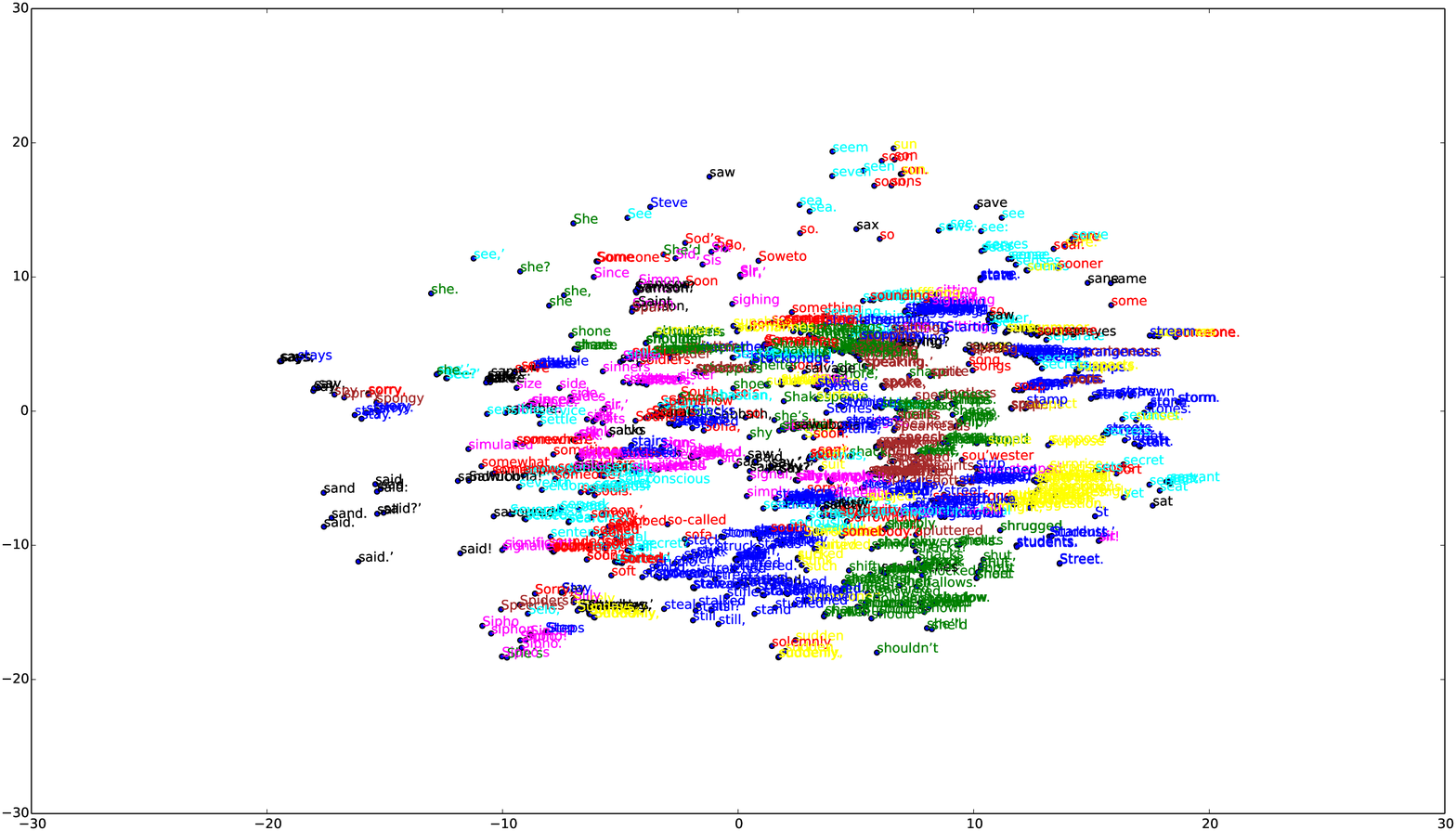}
\caption{TSNE plots for feature representation of unique words starting with S and top eight \textit{second} alphabets ({\color{blue}t}, {\color{darkgreen}h}, {\color{red}o}, {\color{cyan}e}, {\color{magenta}i}, {\color{yellow}u}, {a}, {\color{brown(web)}p}) in word with respect to population in S. (Color based on second alphabet of words starting with S.)}
\label{fig:WordVizLevel2}
\end{subfigure}
\caption{
\textsc{tsne} plots characterizing the quality of feature representations computed by encoder. Each word has a unique high dimensional feature representation which is then embedded in a two dimensional space (using \textsc{tsne}) and is visualized using scatter plots. Similar words are grouped together (shown with various colours) and  dissimilar words tend to get far away. As shown in the figure, (a) words starting with same alphabets belong to same clusters and rest are in other clusters (b) words beginning with 'S' and having same second alphabet belong to same clusters and rest are in other clusters. (Readers are requested to magnify the graphs to look into the intricate details of clusters)}
\label{fig:VisualizationLatentSpace}
\end{figure}

\section{Implementation Details}
Keeping the aspect ratio of input images intact we resize them to height of $30$ pixels. The resized binary images are then used as an input to the two layer \textsc{lstm} encoder-decoder architecture. We use embedding size of $25$ for all our experiments. The dimensionality of output layer in decoder is equal to number of unique symbols in the dataset. \textsc{rms} prop [\cite{rmsProp2012}] with step size of $0.0001$ and momentum of $0.99$ is used to optimize the loss. All relevant parameters are verified and set using a validation set. We use Python's numpy library to implement \textsc{lstm} based architecture. The network is built using computational graphs.

%
 
\section{Experiments}
\footnotetext[1]{Original images are used as input.}  \footnotetext[2]{Images are padded along boundary pixels for better results.}
\footnotetext[3]{Augmented Profiles [\cite{AnandKumarJM07_LSH_Profile}]}

\begin{table}
\begin{center}
\begin{tabular}{|c|c|c|c|}
\hline
\small{\textbf{Features}} & \small{\textbf{Dim}} & \small{\textbf{mAP-100}} & \small{\textbf{mAP-5000}}\\
\hline
\small{\textsc{bow}} & \small{400} & \small{0.5503} & \small{0.33}\\
\hline
\small{\textsc{bow}} & \small{2000} & \small{0.6321} & \small{-}\\
\hline
\small{\textsc{Augmented Profiles\footnotemark[3]}} & \small{247} & \small{0.7371} & \small{0.6189}\\
\hline
\multirow{2}{*}{\small{\textsc{lstm-encoder}}} & \multirow{2}{*}{\small{400}} & \small{0.7402 $(h1-h2)$} & \multirow{2}{*}{\small{0.8521}}\\
& & \small{0.8521 $(c1-c2)$} &\\
\hline
\small{\textsc{ocr - tesseract}} & - & \small{0.6594} & \small{0.7095}\\
\hline
\small{\textsc{ocr - abbyy}} & - & \small{0.8583} & \small{0.872}\\
\hline
\end{tabular}
\end{center}
\caption{mAP computed for various methods: mAP-n stands for mean average precision computed over top n retrievals. $hi$ is hidden representation of layer-$i$ at last timestep of input sequence. $ci$ is memory for layer-$i$ at last timestep of input sequence. A-B is the concatenation of representation A and B. For example, $h1-h2$ represents concatenation of both $h1$ and $h2$.}
\label{tab:retrieval}
\end{table}

\begin{table}
\begin{center}
\begin{tabular}{cc}
\centering
\begin{tabular}{|c|c|}
\hline
\small{\textbf{Model}} & \small{\textbf{Label error(\%)}}\\ 
\hline
\footnotesize{\textsc{abbyy}\footnotemark[2]} & \footnotesize{1.84} \\ \hline
\footnotesize{\textsc{tesseract}\footnotemark[1]} & \footnotesize{35.80}\\ \hline
\footnotesize{\textsc{tesseract}\footnotemark[2]} & \footnotesize{16.95}\\ \hline
\footnotesize{\textsc{rnn Encoder-Decoder}} & \footnotesize{35.57}\\ \hline
\footnotesize{\textsc{lstm-ctc} [\cite{Graves:2006:CTC:1143844.1143891}]} & \footnotesize{0.84}\\ \hline
\footnotesize{\textsc{lstm Encoder-Decoder}} & \footnotesize{0.84}\\ \hline
\end{tabular} &
\begin{tabular}{|c|c|}
\hline
\small{\textbf{Feature}} & \small{\textbf{mAP-100}}\\ \hline
\small{h1-h2}& \small{0.7239}\\ \hline
\small{c1-c2}& \small{0.8548}\\ \hline
\small{h1-h2-c1-c2 L1}& \small{0.8078}\\ \hline
\small{h1-h2-c1-c2 L2}& \small{0.7834}\\ \hline
\small{h1-h2-c1-c2}& \small{0.8545}\\ \hline
\end{tabular}
\end{tabular}
\end{center}
\caption{\textbf{Left:} Label Error Rate comparison of \textsc{rnn}-\textsc{ctc} and Recurrent encoder-decoder.\textbf{Right:} Effect of different concatenation and normalization on features from LSTM-Encoder. \textsc{l1} and \textsc{l2} represent normalization scheme.}
\label{tab:predictions}
\end{table}

\setlength{\tabcolsep}{0.1pt}
\begin{table}[ht!]
\centering
\caption{Qualitative results for retrieval: Comparison of retrieval scheme using simple Bag of Words (BoW) and proposed Deep Word Image Embeddings (DWIE). We use text labels (to save space and more clarity in presentation) of both query and retrieved images to illustrate difference in performance of two approaches. The proposed approach is far more robust to text variations in images and captures much more intricate details about the images.}
\label{tab:qualitative_retrival}
\begin{tabular}{|c|c|c|c|c|c|c|c|c|}
\hline  
& \head{1.5cm}{\textbf{DWIE}}& \head{1.5cm}{\textbf{BoW}}& \head{1.5cm}{\textbf{DWIE}}& \head{1.5cm}{\textbf{BoW}}& \head{1.5cm}{\textbf{DWIE}}& \head{1.5cm}{\textbf{BoW}}& \head{1.5cm}{\textbf{DWIE}}& \head{1.5cm}{\textbf{BoW}} \\ \hline
\begin{tabular}{@{}c@{}}\textbf{Query ($\rightarrow$) /} \\ \textbf{Retrival ($\downarrow$)} \end{tabular}&A.	&A. &following	&following	&returned	&returned	&For 	&For \\ \hline	
R1 &{\color{green} A.}	&{\color{red} "A}	&{\color{green} following}	&{\color{red} long}	&{\color{green} returned}	&{\color{red} turned}	&{\color{green} For}	&{\color{green} For}	\\ \hline
R2 &{\color{green} A.}	&{\color{red} A}	&{\color{green} following}	&{\color{red} long}	&{\color{green} returned}	&{\color{red} read}	&{\color{green} For}	&{\color{green} For}	\\ \hline
R3 &{\color{green} A.}	&{\color{red} A}	&{\color{green} following}	&{\color{red} long}	&{\color{green} returned}	&{\color{red} refused}	&{\color{green} For}	&{\color{green} For}	\\ \hline
R4 &{\color{green} A.}	&{\color{red} At}	&{\color{green} following}	&{\color{green} following}	&{\color{green} returned}	&{\color{red} retorted}	&{\color{green} For}	&{\color{green} For}	\\ \hline
R5 &{\color{green} A.}	&{\color{red} Ah}	&{\color{green} following}	&{\color{red} long}	&{\color{green} returned}	&{\color{red} lifted}	&{\color{green} For}	&{\color{green} For}	\\ \hline
R6&{\color{red} A}	&{\color{red} A}	&{\color{green} following}	&{\color{red} flowing}	&{\color{green} returned}	&{\color{red} ceased}	&{\color{green} For}	&{\color{green} For}	\\ \hline
R7&{\color{red} A}	&{\color{red} A}	&{\color{green} following}	&{\color{red} long}	&{\color{green} returned}	&{\color{red} rolled}	&{\color{green} For}	&{\color{green} For}	\\ \hline
R8&{\color{red} A}	&{\color{red} A}	&{\color{red} folding}	&{\color{red} long}	&{\color{green} returned}	&{\color{red} and}	&{\color{green} For}	&{\color{green} For}	\\ \hline
R9&{\color{red} A}	&{\color{red} A}	&{\color{red} follow}	&{\color{red} long}	&{\color{green} returned}	&{\color{red} carried}	&{\color{green} For}	&{\color{red} for}	\\ \hline
R10&{\color{red} A}	&{\color{red} A}	&{\color{red} followed,}	&{\color{green} following}	&{\color{green} returned}	&{\color{red} red}	&{\color{green} For}	&{\color{red} for}	\\ \hline
R11&{\color{red} A}	&{\color{red} A}	&{\color{red} foolishly}	&{\color{red} long}	&{\color{green} returned}	&{\color{red} Head}	&{\color{green} For}	&{\color{red} for}	\\ \hline
R12&{\color{red} A}	&{\color{red} A}	&{\color{red} fellow,}	&{\color{red} along}	&{\color{green} returned}	&{\color{red} raised}	&{\color{green} For}	&{\color{red} for}	\\ \hline
R13&{\color{red} A}	&{\color{red} A}	&{\color{red} foolscap}	&{\color{red} long}	&{\color{green} returned}	&{\color{red} caused}	&{\color{green} For}	&{\color{red} for}	\\ \hline
R14&{\color{red} A}	&{\color{red} A}	&{\color{red} fellow,}	&{\color{green} following}	&{\color{green} returned}	&{\color{red} turned}	&{\color{green} For}	&{\color{red} for}	\\ \hline
R15&{\color{red} A}	&{\color{red} A}	&{\color{red} foliage.}	&{\color{red} long}	&{\color{green} returned}	&{\color{red} turned}	&{\color{green} For}	&{\color{red} for}	\\ \hline
R16&{\color{red} A}	&{\color{red} A}	&{\color{red} fellow,}	&{\color{red} long}	&{\color{red} returned.}	&{\color{red} road}	&{\color{green} For}	&{\color{red} for}	\\ \hline
R17&{\color{red} A}	&{\color{red} A}	&{\color{red} fellow,}	&{\color{red} long}	&{\color{red} retired}	&{\color{red} and}	&{\color{green} For}	&{\color{red} for}	\\ \hline
R18&{\color{red} A}	&{\color{red} A}	&{\color{red} fellow,}	&{\color{red} long}	&{\color{red} retorted}	&{\color{green} returned}	&{\color{red} Fer-}	&{\color{green} For}	\\ \hline
R19&{\color{red} A}	&{\color{red} As}	&{\color{red} falling}	&{\color{red} long}	&{\color{red} return."}	&{\color{red} God}	&{\color{red} Fer-}	&{\color{red} for}	\\ \hline
R20&{\color{red} A}	&{\color{red} Af-}	&{\color{red} follow,"}	&{\color{red} closing}	&{\color{red} return}	&{\color{red} acted}	&{\color{red} Fer-}	&{\color{green} For}	\\ \hline
\begin{tabular}{@{}c@{}} \textbf{\# relevant matches} \\ \textbf{ in corpus}\end{tabular} & 5	& 5	& 7	& 7	& 15 & 15 & 17 & 17	\\ \hline
\end{tabular}

\end{table}

We demonstrate the utility of the proposed recurrent encoder-decoder framework by two related but independent tasks. Independent baselines are set for both prediction and retrieval experiments.

\textbf{Prediction}: We use $295$K annotated English word images from seven books for our experiments. We perform three way data split for all our experiments -- 60\% training, 20\% validation and remaining 20\% for testing. Results are reported by computing `label error rate'. Label error rate is defined as ratio of sum of insertions, deletions and substitutions relative to length of ground truth over dataset. We compare the results of our pipeline with state-of-art \textsc{lstm-ctc} [\cite{Graves:2006:CTC:1143844.1143891}], an open-source \textsc{ocr tesseract}~[\cite{tesseract_URL}] 
and a commercial \textsc{ocr abbyy}~[\cite{abbyy_URL}].
 
\textbf{Retrieval}: We use $108$K annotated word images from book titled `Adventures of Sherlock Holmes' for retrieval experiments. In all $43$K word images are used for querying the retrieval system. We compare retrieval results with SIFT [\cite{Lowe:2004:DIF:993451.996342}] based bag of words representation, augmented profiles [\cite{AnandKumarJM07_LSH_Profile}] and commercial [\textsc{ocr abbyy}].

\begin{table}
\begin{center}
\centering
\caption{Qualitative results for prediction: We illustrate some of the success and failure cases of our word prediction output. The figure highlights the cases where both commercial and open-source \textsc{ocr}s fail}
\label{tab:qualitative_prediction}
\begin{tabular}{|c|c|c|c|c|}
\hline
\begin{tabular}{@{}c@{}}\small{\textbf{Query Image($\rightarrow$)}}\end{tabular} & \includegraphics[height=4.5mm, trim=0mm 10mm 0mm 10mm, clip]{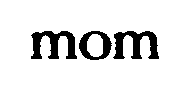}	&\includegraphics[height=4.5mm, trim=0mm 10mm 0mm 10mm, clip]{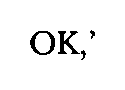} &\includegraphics[height=4.5mm, trim=0mm 10mm 0mm 10mm, clip]{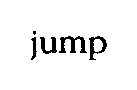}&\includegraphics[height=4.85mm, trim=0mm 10mm 0mm 10mm, clip]{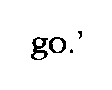} \\ \hline	
\small{True Label}&\small{mom} &\small{OK,'} &\small{jump} &\small{go.'}\\ \hline	
\small{Tessaract~\footnotemark[2]} &\small{IDOITI} &\small{ox,'} &\small{iump} &\small{80.} \\ \hline	
\small{Abbyy~\footnotemark[2]}  &\small{UJOUJ} &\small{ok;} &\small{duinl} &\small{g�-'} \\ \hline	
\small{LSTM-ED}&\small{mom} &\small{OK,'} &\small{jump} &\small{go.'}\\ \hline	
\end{tabular}
\end{center}
\end{table}

\subsection{Results and Discussion}

Table \ref{tab:predictions} exhibits prediction baseline. We observe that \textsc{lstm Encoder-Decoder} outperforms vanilla \textsc{rnn Encoder-Decoder} by significant margin. It even scores better when compared to \textsc{lstm} with \textsc{ctc} output layer and \textsc{abbyy}. When compared to \textsc{ctc} layer based \textsc{lstm} networks, our network requires more memory space. The strength of our network is fixed length representation for variable length input which enables us to perform better and faster retrieval. 

Table \ref{tab:retrieval} depicts retrieval baseline. Features from \textsc{lstm} encoder (referred as deep word image embedding (\textsc{dwie})) are used for comparisons with other state-of-art results. We observe that \textsc{dwie} features significantly outperform \textsc{sift} [\cite{Lowe:2004:DIF:993451.996342}] based bag of words (\textsc{bow}) and augmented profiles [\cite{AnandKumarJM07_LSH_Profile}]. When compared to \textsc{abbyy}, \textsc{dwie} features perform a notch better for top $5000$ retrieval but perform similar for top $100$ retrieval. The memory states at last position of each sequence are used as \textsc{dwie} features. Various normalization (L1 and L2) and augmentations with hidden states were tried out as shown in Table \ref{tab:predictions}. 

Table \ref{tab:qualitative_retrival} demonstrates the qualitative performance of retrieval system using both deep word image embedding (\textsc{dwie}) and bag of words (\textsc{bow}) models. The table illustrates top $20$ retrievals using both the methods. We observe the proposed embeddings to be better than naive (\textsc{bow}) in such settings. In majority of the cases we find all relevant(exact) matches at top in case of deep embeddings, which is not the case with (\textsc{bow}) model.  \textsc{dwie} seems highly sensitive to small images components like `.' (for query `A.') which is not the case with \textsc{bow} model. Simple \textsc{bow} fails to recover any relevant samples for query `A.' in top $20$ retrievals. 

Figure \ref{fig:VisualizationLatentSpace} shows \textsc{t-sne} [\cite{tSNE}] plots of word image encodings. We show two levels of visualization along with groupings in context of word image representation. It's clear from the figure \ref{fig:WordVizLevel1} that representation is dominated by first character of the word in word image. Sequence of correct encodings play a major role in full word prediction -- a wrong letter prediction in early stages would result in overall invalid word prediction. 

%

\begin{figure}
\begin{subfigure}{.5\textwidth}
  \centering
\includegraphics[height=5cm, trim=20mm 10mm 30mm 20mm, clip]{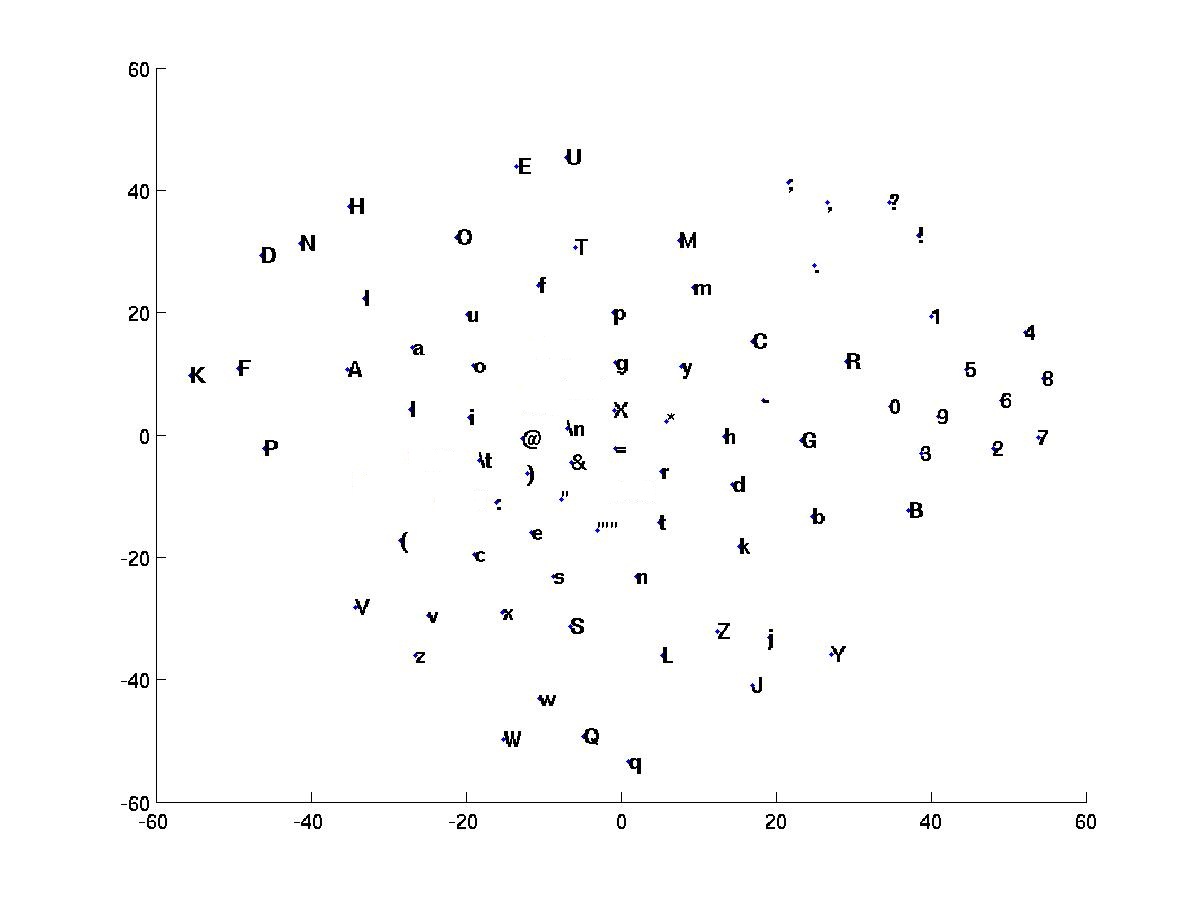}
\caption{Learnt embedding of characters using t\textsc{sne}.}
\label{fig:charEmbedding}
\end{subfigure}%
\begin{subfigure}{.5\textwidth}
  \centering
\includegraphics[height=5cm, trim=10mm 5mm 10mm 10mm, clip]{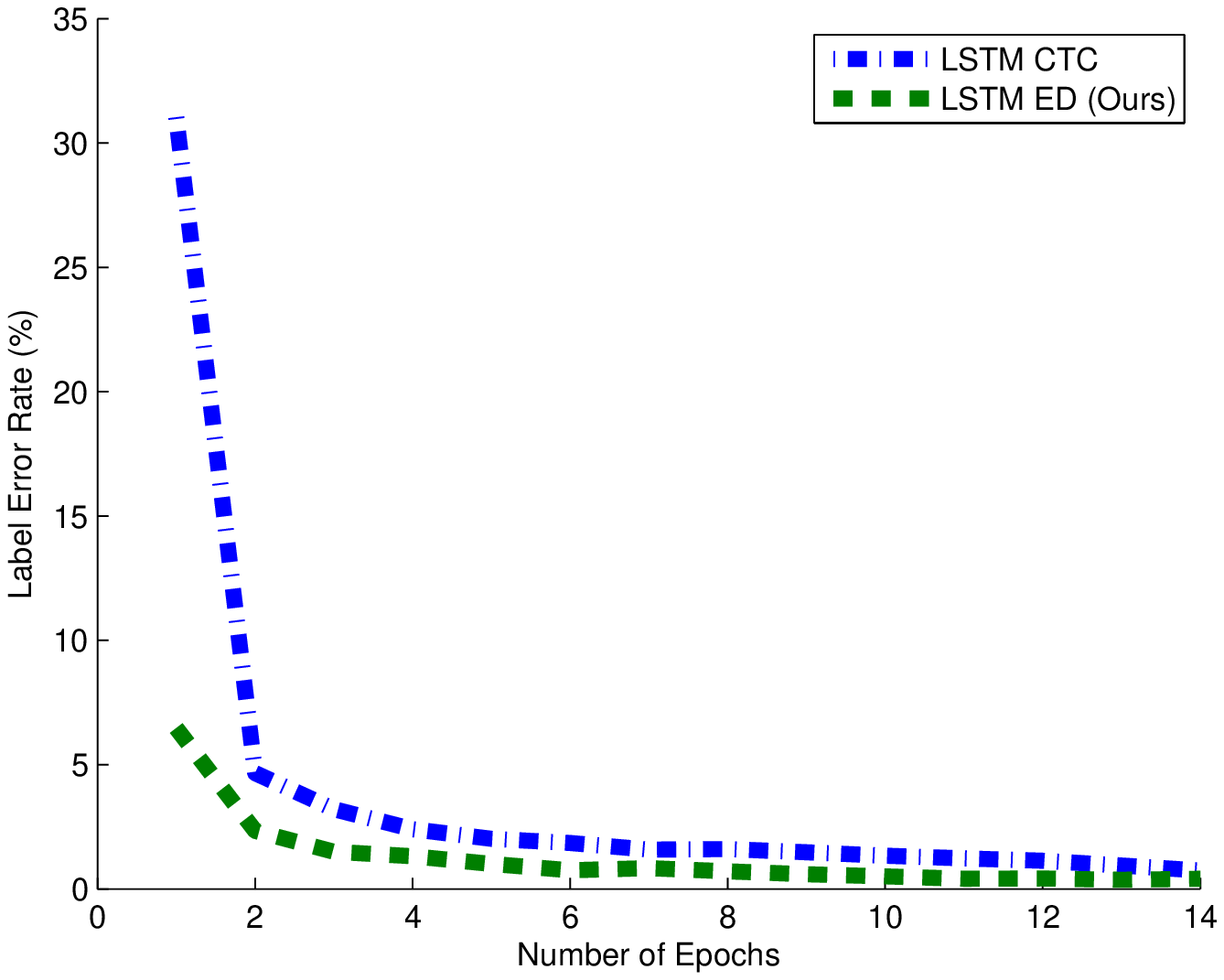}
\caption{Convergence of \textsc{lstm-ctc} \protect\footnotemark[4] and \textsc{lstm-ed}.}
\label{fig:convergence}
\end{subfigure}%
\caption{Plot showing character embedding and convergence of \textsc{lstm-ctc} \protect\footnotemark[4] and \textsc{lstm-ed}.}
\end{figure}

Figure \ref{fig:charEmbedding} is a plot of learnt embeddings which shows relative similarities of characters. The similarities are both due to structure and language of characters -- (i) all the numbers (0-9) are clustered together (ii) punctuations are clustered at top right of graph (iii) upper case and lower case characters tend to cluster together, viz. (m,M), (v,V), (a,A) etc. As embeddings are learnt jointly while minimizing cost for correct predictions, they tend to show relative similarity among nearby characters based jointly on structure in image space and language in output space. Figure \ref{fig:convergence} illustrates training label error rate for various learning models -- \textsc{lstm} with \textsc{ctc} output layer and \textsc{lstm} encoder-decoder.

\footnotetext[4]{Standard LSTM-CTC implementation [\cite{Graves:2006:CTC:1143844.1143891}].}  

\section{Conclusion}
We demonstrate the applicability of sequence to sequence learning for word prediction in printed text \textsc{ocr}. Fixed length representation for variable length input using a recurrent encoder-decoder architecture sets us apart from present day state of the art algorithms. We believe with enough memory space availability, sequence to sequence regime could be a better and efficient alternative for \textsc{ctc} based networks. The network could well be extended for other deep recurrent architectures with variable length inputs, e.g.\ attention based model to describe the image contents etc.\

{\small
\bibliography{egbib_v2}
\bibliographystyle{iclr2015}
}
\end{document}